\documentclass[journal]{IEEEtran}

\usepackage{graphicx}
\usepackage{xcolor}
\usepackage{booktabs}
\usepackage{multirow}
\usepackage[detect-none]{siunitx}
\usepackage[export]{adjustbox}
\usepackage{amssymb}
\usepackage{amsmath}
\usepackage{longtable}
\usepackage{pifont}
\usepackage{placeins}
\usepackage[hidelinks]{hyperref}

\graphicspath{{images}}
\newcommand{\ie}{{i.e.},~}
\newcommand{\eg}{{e.g.},~}

\DeclareSIUnit{\nothing}{\relax}
\DeclareSIUnit{\mac}{MAC}
\DeclareSIUnit{\frame}{frame}
\DeclareSIUnit{\pixel}{px}
\DeclareSIUnit{\rad}{rad}

\newcommand{\relpose}[2]{{\mathbf{T}_{#1}^{#2}}} %\;
\newcommand{\ident}{{\mathbf{I}}}

\newcommand{\cmark}{\ding{51}}%
\newcommand{\xmark}{\ding{55}}%

\newcommand{\trainall}{\textit{all}}
\newcommand{\trainfc}{\textit{fc}}
\newcommand{\trainbn}{\textit{bn}}
\newcommand{\trainbias}{\textit{bias}}

\newcommand{\lossid}{Supervised}
\newcommand{\lossre}{SSL}

\makeatletter
\newcommand{\todo}[1]{\noindent\textit{\color{red}\textbf{TODO}~#1}\@latex@warning{TODO: #1}}%LS
\makeatother

\newcommand{\rebuttal}[1]{\noindent#1}% LS
\newcommand{\rebuttaltable}{}% LS
\newadjustimage{\rebuttalfigure}[2][center]{width={#2},{#1}}% LS

\begin{document}

\title{Training on the Fly: On-device Self-supervised Learning aboard Nano-drones within \SI{20}{\milli\watt}}

% \author{Authors omitted for review}

\author{Elia~Cereda,~\IEEEmembership{Student Member,~IEEE,}
        Alessandro~Giusti,~\IEEEmembership{Member,~IEEE,}
        and~Daniele~Palossi% <-this % stops a space
\thanks{Manuscript received March 31, 2024; revised June 16, 2024.}
\thanks{This work has been partially funded by the Hasler Foundation (Grant \# 23059).
The authors would like to thank Mirko Nava, Davide Nadalini, and Manuele Rusci for their support and useful discussions.}% <-this % stops a space
\thanks{E. Cereda, A. Giusti, and D. Palossi are with the Dalle Molle Institute for Artificial Intelligence~(IDSIA), USI-SUPSI, 6962 Lugano, Switzerland. Contact author: {\tt\small elia.cereda@idsia.ch}.}% <-this % stops a space
\thanks{D. Palossi is also with the Integrated Systems Laboratory (IIS), ETH Z\"urich, 8092 Z\"urich, Switzerland.}% <-this % stops a space
}

% \markboth{IEEE Transactions On Computer-aided Design of Integrated Circuits and Systems,~Vol.~\#, No.~\#, \#\#~2024}{}% {Cereda \MakeLowercase{\textit{et al.}}: \thetitle}
% \fi


% \IEEEspecialpapernotice{(Invited Paper)}

\maketitle

\begin{abstract}
Miniaturized cyber-physical systems (CPSes) powered by tiny machine learning (TinyML), such as nano-drones, are becoming an increasingly attractive technology.
Their small form factor (\ie, $\sim$\SI{10}{\centi\meter} diameter) ensures vast applicability, ranging from the exploration of narrow disaster scenarios to safe human-robot interaction.
Simple electronics make these CPSes inexpensive, but strongly limit the computational, memory, and sensing resources available on board.
\rebuttal{In real-world applications, these limitations are further exacerbated by domain shift.
This fundamental machine learning problem implies that model perception performance drops when moving from the training domain to a different deployment one.}
To cope with and mitigate this general problem, we present a novel on-device fine-tuning approach that relies only on the limited ultra-low power resources available aboard nano-drones.
Then, to overcome the lack of ground-truth training labels aboard our CPS, we also employ a self-supervised method based on ego-motion consistency.
Albeit our work builds on top of a specific real-world vision-based human pose estimation task, it is widely applicable for many embedded TinyML use cases.
\rebuttal{Our 512-image on-device training procedure is fully deployed aboard an ultra-low power GWT GAP9 System-on-Chip and requires only \SI{1}{\mega\byte} of memory while consuming as low as \SI{19}{\milli\watt} or running in just \SI{510}{\milli\second} (at \SI{38}{\milli\watt}).}
Finally, we demonstrate the benefits of our on-device learning approach by field-testing our closed-loop CPS, showing a reduction in horizontal position error of up to 26\% vs. a non-fine-tuned state-of-the-art baseline.
In the most challenging never-seen-before environment, our on-device learning procedure makes the difference between succeeding or failing the mission.
\end{abstract}

\begin{IEEEkeywords}
On-device Learning, Self-supervised Learning, Embedded ML, TinyML, Resource-constrained CPS.
\end{IEEEkeywords}

\section*{Supplementary video material}
Supplementary video material of the in-field experiments at \url{https://youtu.be/3yNbMwszpSY}

% \fi
\IEEEpeerreviewmaketitle

\section{Introduction} \label{sec:introduction}

\IEEEPARstart{M}{iniaturized} unmanned aerial vehicles (UAVs) as small as the palm of one hand, also known as nano-UAVs, are appealing cyber-physical systems (CPS), which, leveraging tiny machine learning (TinyML) algorithms deployed aboard, have reached an unprecedented level of autonomy.
Thanks to their small form factor, i.e., \SI{10}{\centi\meter} in diameter and sub-\SI{50}{\gram} in weight, nano-UAVs can fly in narrow and constrained spaces~\cite{palossi201964} or safely in the human proximity~\cite{pulp-frontnet} embodying the ultimate dynamic IoT smart sensors, capable of analyzing their surrounding and flying where their presence is most needed.
To cope with extremely limited onboard sensory, memory, and computational resources, \ie, low-resolution cameras, a few \SI{}{\mega\byte} off-chip memories, and sub-\SI{100}{\milli\watt} power envelope for the computational units, recent works make extensive use of optimized TinyML workloads, such as convolutional neural networks (CNNs)~\cite{pulp-frontnet, nanoflownet, lamberti22tinydronet}.

\rebuttal{Apart from the well-known challenges in deploying complex deep learning (DL) models on ultra-low-power microcontroller units (MCUs), TinyML algorithms are vulnerable to the fundamental problem of \textit{domain shift}~\cite{9782500}.
Domain shift occurs when a DL model, such as a classifier or regressor, trained on data acquired in a given context, i.e., \textit{domain}, is deployed in a different one.}
Then, the predictive performance of the system (i.e., the accuracy or estimation error) often decreases because the training data is not representative of the deployment domain. 
This problem is particularly present in real-world applications, such as the robotic task we address in this work.
\textbf{This paper explores on-device learning as a viable and effective solution to the domain shift problem.}

We consider the case in which a CNN, previously trained on a given task using large datasets, is \textit{fine-tuned} on-device, using a small amount of data collected after deployment.
Fine-tuning consists of updating the parameters of the pre-trained models by executing a limited number of additional training steps.
With on-device self-supervised learning, data can be collected by the robot precisely in the domain in which the model will operate, counteracting the domain shift.
Despite on-device learning being conceptually simple and attractive, pulling it off in a real-world application is extremely challenging.
Fine-tuning typically requires significant memory and computational resources, including hardware support for floating-point arithmetic (not always available on MCUs) and data availability with the corresponding ground-truth labels.

\rebuttal{
In this work, we study this problem in the context of a real-world robotic application, covering the full pipeline of design, deployment, and testing of a vertically integrated on-device learning system and evaluating the practical implications of our design choices.
In particular, we strive to answer three key research questions:
\textit{i)} What is the best fine-tuning strategy to exploit the limited computing power?
\textit{iii)} Can the limited on-board memory fit enough data for fine-tuning to be effective?
\textit{ii)} How to deal with the lack of ground-truth data in the field?
While analysed in a specific context, these questions yield valuable insights for real-world applications of on-device learning also on other tasks.
}

\begin{table*}
\centering
\caption{Embedded on-device learning literature review. We report the lowest power computational device considered in each work.}
\label{tab:related-work}
\renewcommand{\arraystretch}{0.9}
\begin{tabular}{cccccccccc}
\toprule
    \multirow{2}{*}{\textbf{Work}} & 
    \multirow{2}{*}{\textbf{Task}} & 
    \multirow{2}{*}{\textbf{Arch.}} & 
    \multirow{2}{*}{\textbf{Data}} & 
    \multirow{2}{*}{\textbf{Training}} & 
    \multirow{2}{*}{\textbf{Supervision}} & 
    \multirow{2}{*}{\textbf{Compute}} & 
    \multirow{2}{*}{\shortstack[c]{\textbf{Memory}\\{\textbf{[\si{\mega\byte}]}}}} & 
    \multirow{2}{*}{\shortstack[c]{\textbf{Power}\\{\textbf{[\si{\watt}]}}}} & 
    \multirow{2}{*}{\shortstack[c]{\textbf{Field-}\\\textbf{tested}}} \\
    \\
\midrule
    \multirow{2}{*}{Mudrakarta \textit{et al.}~\cite{mudrakarta2018k}} & \multirow{2}{*}{\shortstack[c]{object class. +\\object detect.}} & \multirow{2}{*}{CNN} & \multirow{2}{*}{\shortstack[c]{benchmark\\datasets}} & \multirow{2}{*}{\shortstack[c]{fine-\\tuning}} & \multirow{2}{*}{\shortstack[c]{supervised}} & \multirow{2}{*}{\shortstack[c]{GPU\\(unspecified)}} & \multirow{2}{*}{1000s} & \multirow{2}{*}{100s} & \multirow{2}{*}{\xmark} \\
    \\
    \rule{0pt}{3.25ex}
    \multirow{2}{*}{TinyTL~\cite{cai2020tinytl}} & \multirow{2}{*}{\shortstack[c]{object\\classification}} & \multirow{2}{*}{CNN} & \multirow{2}{*}{\shortstack[c]{benchmark\\datasets}} & \multirow{2}{*}{\shortstack[c]{fine-\\tuning}} & \multirow{2}{*}{\shortstack[c]{supervised}} & \multirow{2}{*}{\shortstack[c]{GPU\\(unspecified)}} & \multirow{2}{*}{1000s} & \multirow{2}{*}{100s} & \multirow{2}{*}{\xmark} \\
    \\
    \rule{0pt}{3.25ex}
    \multirow{2}{*}{PULP-TrainLib~\cite{NADALINI2023212}} & \multirow{2}{*}{\shortstack[c]{object class. +\\keyword spotting}} & \multirow{2}{*}{CNN} & \multirow{2}{*}{\shortstack[c]{benchmark\\datasets}} & \multirow{2}{*}{\shortstack[c]{from\\scratch}} & \multirow{2}{*}{\shortstack[c]{supervised}} & \multirow{2}{*}{\shortstack[c]{GWT\\GAP9}} & \multirow{2}{*}{0.4} & \multirow{2}{*}{0.1} & \multirow{2}{*}{\xmark} \\ % 05.2023
    \\
    \rule{0pt}{3.25ex}
    \multirow{2}{*}{RLtools~\cite{eschmann2023rltools}} & \multirow{2}{*}{\shortstack[c]{robot control}} & \multirow{2}{*}{\shortstack[c]{fully\\connected}} & \multirow{2}{*}{\shortstack[c]{external\\simulator}} & \multirow{2}{*}{\shortstack[c]{from\\scratch}} & \multirow{2}{*}{\shortstack[c]{reinforcement}} & \multirow{2}{*}{\shortstack[c]{i.MX\\RT1060}} & \multirow{2}{*}{$16$} & \multirow{2}{*}{0.3} & \multirow{2}{*}{\xmark} \\ % 06.2023
    \\
    \rule{0pt}{3.25ex}
    \multirow{2}{*}{TinyTrain~\cite{kwon2023tinytrain}} & \multirow{2}{*}{\shortstack[c]{object\\classification}} & \multirow{2}{*}{CNN} & \multirow{2}{*}{\shortstack[c]{benchmark\\datasets}} & \multirow{2}{*}{\shortstack[c]{fine-\\tuning}} & \multirow{2}{*}{\shortstack[c]{supervised}} & \multirow{2}{*}{\shortstack[c]{Raspberry\\Pi Zero 2}} & \multirow{2}{*}{$512$} & \multirow{2}{*}{10} & \multirow{2}{*}{\xmark} \\ % 07.2023
    \\
    \rule{0pt}{3.25ex}
    \multirow{2}{*}{PockEngine~\cite{zhu23pockengine}} & \multirow{2}{*}{\shortstack[c]{object\\classification}} & \multirow{2}{*}{\shortstack[c]{CNN +\\Transformer}} & \multirow{2}{*}{\shortstack[c]{benchmark\\datasets}} & \multirow{2}{*}{\shortstack[c]{fine-\\tuning}} & \multirow{2}{*}{\shortstack[c]{supervised}} & \multirow{2}{*}{STM32F7} & \multirow{2}{*}{0.3} & \multirow{2}{*}{0.2} & \multirow{2}{*}{\xmark} \\ % 10.2023
    % Power consumption 1082 CoreMark / 6 CoreMark/mW = 180mW, from https://www.st.com/resource/en/brochure/brstm32f7.pdf
    \\
    \rule{0pt}{3.25ex}
    \multirow{2}{*}{LifeLearner~\cite{kwon2023lifelearner}} & \multirow{2}{*}{\shortstack[c]{object\\classification}} & \multirow{2}{*}{\shortstack[c]{CNN}} & \multirow{2}{*}{\shortstack[c]{benchmark\\datasets}} & \multirow{2}{*}{\shortstack[c]{continual\\learning}} & \multirow{2}{*}{\shortstack[c]{supervised}} & \multirow{2}{*}{STM32H7} & \multirow{2}{*}{1} & \multirow{2}{*}{1} & \multirow{2}{*}{\xmark} \\ % 11.2023
    % Power consumption 620mA * 1.8V = 1.1W from Table 20 and 21 of https://www.st.com/resource/en/datasheet/stm32h747ag.pdf
    \\
    \rule{0pt}{3.25ex}
    \multirow{2}{*}{\textbf{Ours}} & \multirow{2}{*}{\shortstack[c]{\textbf{human pose}\\\textbf{estimation}}} & \multirow{2}{*}{\textbf{CNN}} & \multirow{2}{*}{\shortstack[c]{\textbf{real-world}\\\textbf{robot}}} & \multirow{2}{*}{\shortstack[c]{\textbf{fine-}\\\textbf{tuning}}} & \multirow{2}{*}{\shortstack[c]{\textbf{self-supervised}}} & \multirow{2}{*}{\shortstack[c]{\textbf{GWT}\\\textbf{GAP9}}} & \multirow{2}{*}{\textbf{1}} & \multirow{2}{*}{\textbf{0.1}} & \multirow{2}{*}{\cmark} \\
    \rule{0pt}{2ex}
    \\
\bottomrule
\end{tabular}
\end{table*}

\rebuttal{We start from the State-of-the-Art (SotA) PULP-Frontnet CNN~\cite{pulp-frontnet} for the human pose estimation task, whose outputs control the nano-UAV and allow to keep it in front of the user at a predefined distance.}
We tackle the first computational challenges by adopting an ultra-low-power GWT GAP8 System-on-Chip (SoC) and its next generation, the GAP9.
Then, we investigate four fine-tuning strategies, spanning from the most memory-hungry fine-tuning of all layers of the CNN down to minimal fine-tuning of only the final fully connected layer.
Further, we propose a self-supervised state-consistency loss term~\cite{nava2021stateconsist} to handle the data availability problem, relieving our fine-tuning process from ground-truth labels.
Our \textbf{main contributions} are the following:
\begin{itemize}
    \item an in-depth analysis of four fine-tuning strategies, which explores the trade-off between computational and memory requirements and their prediction performance in a real-world scenario. Our results show a regression performance improvement up to 30\% of our self-supervised method against a non-fine-tuned baseline, which grows to 56\% when using ground-truth labels.
    \item \rebuttal{On-device implementation and profiling of the resulting best fine-tuning approach, requiring only \SI{6.6}{\second}@\SI{102}{\milli\watt} and \SI{511}{\milli\second}@\SI{38}{\milli\watt} for fine-tuning on 512 images (5 epochs) on the GAP8 and GAP9 SoC, respectively.}
    \item \rebuttal{An in-field evaluation in a very challenging deployment field where the SotA} PULP-Frontnet baseline fails in following the user, while all self-supervised fine-tuned models can complete between 92 and 100\% of the expected path (over 3 runs for each model).
    \item A quantitative evaluation of the nano-UAV position error resulting from an autonomous closed-loop controller that uses the fine-tuned model for perception. Compared to a pre-trained perception model, it reduces by 26\% the horizontal position error. 
\end{itemize}
Our work marks the first real-world demonstration of on-device learning aboard a nano-UAV, addressing the critical domain shift problem and paving the way for more general advancements in the scientific community.
\section{Related work}
\label{sec:related-work}

Domain shift~\cite{9782500} is a significant challenge for machine learning approaches in every field: models that perform well on their training domain often underperform when deployed because real-world conditions differ or even change over time.
Robotics further exacerbates the issue due to a scarcity of training data, which compounds on hardware-constrained platforms such as nano-UAVs, with the tiny DNN architectures affordable aboard.
Past efforts address this issue by better-taking advantage of the limited real-world training data~\cite{cereda2022pitchaug}, generating vast training datasets in simulation~\cite{moldagalieva23dataset}, and taking advantage of additional sensors that are more robust to sim-to-real domain shift (e.g., depth sensors)~\cite{crupi2023simtoreal}.

On-device learning is an emerging field that proposes a radically different approach~\cite{pavan2023,ren2021tinyol,pau2023}: abandon the traditional \textit{train-once, deploy-everywhere} paradigm by enabling devices to adapt a model to their domain directly in the field.
Table~\ref{tab:related-work} gives an overview of the most significant on-device learning literature in chronological order.
However, applying on-device learning to nano-UAVs is still hurdled by the limitations of the onboard sensors and computationally-constrained processors.
Compared to larger-scale drones~\cite{decroon_alphapilot}, nano-UAVs can only afford MCU devices with 1/1000th the memory and computational power of their bigger counterparts. Embodiments of the Parallel Ultra Low Power (PULP) platform~\cite{rossi2021vega}, in particular, have enabled a number of break-through applications on nano-UAVs~\cite{nanoflownet,lamberti22tinydronet}.
Regarding on-device learning on PULP devices, \textit{PULP-TrainLib}~\cite{NADALINI2023212} was recently proposed to accelerate neural network back-propagation through parallelization, software optimizations, and reduced 16-bit floating-point precision.

A number of approaches have been also demonstrated on other embedded platforms.
For example, gradient rescaling, sparse updates, and compile-time graph optimization allow \textit{PockEngine}~\cite{zhu23pockengine} to achieve 8-bit approximated backward passes on STM32 MCUs.
\textit{RLtools}~\cite{eschmann2023rltools} introduce a highly portable on-device reinforcement learning implementation and test it, among others, on an i.MX RT1060 MCU.
Despite being tailored for robotics applications, RLtools focus on fully connected models and control tasks, while our focus is on more computationally intensive perception tasks with CNN-based methods.

\rebuttal{Various approaches propose to reduce the re-training workload through \textit{spare updates}, \ie by fine-tuning only a subset of the model parameters~\cite{mudrakarta2018k, cai2020tinytl, kwon2023tinytrain, kwon2023lifelearner}.}
Mudrakarta \textit{et al.}~\cite{mudrakarta2018k} limits updates to just the batch normalization layers, which TinyTL~\cite{cai2020tinytl} further reduces to only the bias parameters.
The latter translates to significant memory savings (up to 6.5$\times$) by not storing activation maps during the forward pass since they are needed only for back-propagation to weight parameters, not biases.
TinyTrain~\cite{kwon2023tinytrain} introduces a dynamic strategy based on task-adaptive sparse updates, which achieves a 5.0\% accuracy gain on image classification benchmarks and reduces the backward-pass memory and computation by up to 2.3$\times$ and 7.7$\times$, respectively.
LifeLearner~\cite{kwon2023lifelearner} applies lossy compression to the stored activation maps to reduce their size in memory by at least 11.4$\times$.
\rebuttal{Crucially, these methods are tested only on image classification tasks and assume that a training dataset with ground-truth labels is provided externally.
Further, not all methods are deployed and evaluated on the target embedded devices.
On the contrary, our approach extends these techniques~\cite{mudrakarta2018k,cai2020tinytl} to a regression task, and deploys them on-device and in the real world, where ground-truth labels are not readily available.
Addressing these fundamental problems has a far-reaching impact on many real-life perception problems.}

\begin{figure}[t]
  \centering
  \includegraphics[width=1.01\columnwidth]{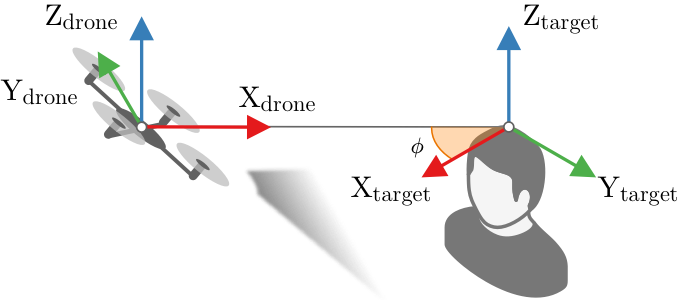}
  \caption{Human pose estimation task and reference frames~\cite{cereda2022pitchaug}.}
  \label{fig:task}
\end{figure}

Exploiting self-collected data and deriving supervisory information exclusively from onboard sensors (as opposed to external infrastructure, manual annotation, or user intervention) is the core idea of self-supervised learning.
Three categories of approaches have been recently applied in autonomous robotics.
Most approaches derive noisy approximations to the ground-truth labels using task-specific methods, such as exploring an environment with a drone until it crashes~\cite{gandhi17flybycrashing} or while continuously measuring its distance from the surrounding environment~\cite{kouris18flybymyself}.
Others learn a secondary task, for which ground-truth labels are known to the robot, as a pretext for learning the task of interest, for which no or little data is available. 
Namely, predicting a quadrotor's sound from camera images is a strong proxy for estimating its location~\cite{nava22soundpretext}.
Additionally, a masked autoencoder for image reconstruction can be adapted to perform a range of robot manipulation tasks~\cite{radosavovic2022}.
Finally, audio, optical flow, and depth estimation are useful pretexts for visual odometry~\cite{lai23xvo}.
The third group imposes geometric consistency constraints to improve model predictions: an object pose estimation model must be consistent over time with the robot's ego motion~\cite{nava2021uncertainty}, while visual odometry between pairs of camera images must satisfy the transitive property~\cite{iyer18geomconsistency}.
The photometric reprojection error is another common constraint used to train self-supervised monocular depth~\cite{godard19depth}, optical flow~\cite{liu19selflow}, and visual odometry~\cite{yang20d3vo,bian19unsupervised} models.

\begin{figure}[t]
  \centering
  \includegraphics[width=0.65\columnwidth]{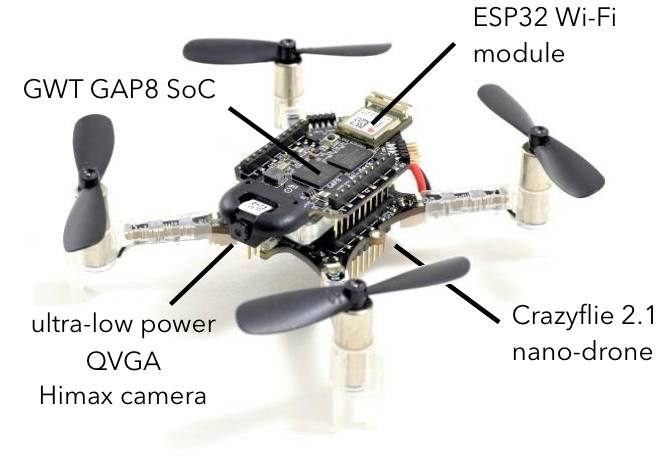}
  \caption{Robot platform: Crazyflie 2.1 with AI-deck and Flow-deck boards.}
  \label{fig:crazyflie}
\end{figure}

Our work proposes a self-supervised fine-tuning process based on an ego-motion consistency loss and approximated labels derived through task-specific collaboration with the user.
However, in contrast to the above approaches, which assume abundant self-supervised data, we are tightly constrained by the amount of fine-tuning data we can store on our embedded system's 8MB DRAM memory.
Additionally, previous consistency approaches assume self-supervised labels that are noisy but fully measurable (e.g., learning object pose estimation in the presence of odometry error while the object remains still~\cite{nava2021uncertainty}). 
By comparison, we also deal with inherently unknown data, such as the movements of human subjects.
\section{Background}
\label{sec:background}

\subsection{Regression task}
We demonstrate our approach aboard an autonomous nano-UAV that performs the human pose estimation task~\cite{pulp-frontnet}.
A perception CNN takes gray-scale $160\times\SI{96}{\pixel}$ camera frames and estimates the subject's 4DOF pose $[x, y, z, \phi]$, relative to the drone's horizontal frame as shown in Figure~\ref{fig:task}. 
Therefore, the model output poses do not depend on the pitch and roll orientations of the drone and human subjects.
The poses are processed with a Kalman filter and fed to a closed-loop velocity controller that autonomously flies the drone, keeping a fixed distance $\Delta$ in front of the human subject and centering them in the image.

\newcommand{\tablecell}[1]{\multirow{2}{*}{\normalsize{#1}}}%
\begin{table}[t]
    \centering
    \caption{Comparison of SoCs aboard Crazyflie nano-UAVs}
    \label{tab:soc_compare}
    \resizebox{\columnwidth}{!}{
        \renewcommand{\arraystretch}{1.2}
        \begin{tabular}{ccccccccc}
        \toprule
        \multirow{2}{*}{\shortstack[c]{\textbf{SoC}\\(ISA)}} &
        \multirow{2}{*}{\shortstack[c]{\textbf{CPU}\\\textbf{cores}}} & 
        \multirow{2}{*}{\shortstack[c]{\textbf{L1}\\\textbf{[\si{\kilo\byte}]}}} & 
        \multirow{2}{*}{\shortstack[c]{\textbf{L2}\\\textbf{[\si{\mega\byte}]}}} & 
        \multirow{2}{*}{\shortstack[c]{\textbf{L3}\\\textbf{[\si{\mega\byte}]}}} & 
        \multirow{2}{*}{\shortstack[c]{\textbf{SIMD}\\\textbf{instr.}}} &
        \multirow{2}{*}{\textbf{FPUs}} &
        \multirow{2}{*}{\shortstack[c]{\textbf{Max freq.}\\\textbf{[\si{\mega\hertz}]}}} &
        \multirow{2}{*}{\shortstack[c]{\textbf{Power}\\\textbf{[\si{\milli\watt}]}}} \\
        \\
        \midrule
        \multirow{2}{*}{\shortstack[c]{\textbf{STM32}\\(ARM)}}    & \tablecell{1}   & \tablecell{64} & \tablecell{0.1} & \tablecell{--} & \tablecell{--} &      \tablecell{1}
                                                                  & \tablecell{168} & \tablecell{116} \\ 
        \\
        \multirow{2}{*}{\shortstack[c]{\textbf{ESP32}\\(Xtensa)}} & \tablecell{2}   & \tablecell{32} & \tablecell{0.5} & \tablecell{--}
                                                                  & \tablecell{--}  & \tablecell{2}
                                                                  & \tablecell{240} & \tablecell{300} \\ 
        \\
        \multirow{2}{*}{\shortstack[c]{\textbf{GAP8}\\(RISC-V)}}  & \multirow{2}{*}{\shortstack[c]{8\\(+1)}}       &  \tablecell{64} & \tablecell{0.5} & \tablecell{8}
                                                                  & \multirow{2}{*}{\cmark} &      \tablecell{--}
                                                                  & \tablecell{175} & \tablecell{96} \\ 
        \\
        \multirow{2}{*}{\shortstack[c]{\textbf{GAP9}\\(RISC-V)}}  & \multirow{2}{*}{\shortstack[c]{\textbf{9}\\\textbf{(+1)}}}       &  \tablecell{\textbf{128}} & \tablecell{\textbf{1.5}} & \tablecell{\textbf{32}} 
                                                                  & \multirow{2}{*}{\cmark} &      \tablecell{\textbf{4}}
                                                                  & \tablecell{\textbf{370}} & \tablecell{\textbf{66}} \\ 
        \\
        \bottomrule
        \end{tabular}
    }
\end{table}

\subsection{Robotic platform}
Our target robot is the Bitcraze Crazyflie 2.1, a commercial off-the-shelf (COTS) nano-UAV, extended by the pluggable AI-deck and Flow-deck companion boards as depicted in Figure~\ref{fig:crazyflie}.
Table~\ref{tab:soc_compare} compares the computational and memory resources available on this drone. 
The Crazyflie relies on an STM32 single-core microcontroller unit (MCU) for low-level flight control and state estimation. It can reach up to \SI{7}{\minute} flight time on a single \SI{380}{\milli\ampere\hour} battery.
The AI-deck extends onboard sensing and processing capabilities with a GreenWaves Technologies (GWT) GAP8 SoC and a Himax HM01B0 gray-scale QVGA camera, while the Flow-deck provides a time-of-flight laser-based altitude sensor and an optical flow sensor to improve the drone's state estimation for autonomous flight.

The GAP8 SoC is an embodiment of the parallel ultra-low power platform (PULP)~\cite{rossi2021vega}, shown in Figure~\ref{fig:pulp_architecture}.
It features two power domains: a computationally-capable 8-core cluster (CL) and a single-core fabric controller (FC), in charge of data orchestration for the CL's execution.
All cores are based on the RISC-V instruction set architecture; the FC can reach up to \SI{250}{\mega\hertz}, while the CL peaks at \SI{175}{\mega\hertz}.
The on-chip memories are organized in a fast \SI{64}{\kilo\byte} scratchpad L1 and a slower \SI{512}{\kilo\byte} L2 memory.
Additionally, the AI-deck features also \SI{8}{\mega\byte} off-chip DRAM memory and \SI{64}{\mega\byte} Flash.
Finally, the GAP8 does not provide any hardware support for floating-point calculations and requires either costly soft-float emulation ($10\times$ overhead) or fixed-point arithmetic through quantization.

The next-generation SoC GAP9 marks significant improvements compared to GAP8, as shown in Table~\ref{tab:soc_compare}. 
Most significantly, GAP9's CL includes four shared floating point units (FPUs), which execute floating-point instructions in a single clock cycle. 
In the context of on-device training, floating-point hardware support is extremely valuable for back-propagation, for which the basic primitives are implemented in the \textit{PULP-TrainLib}~\cite{NADALINI2023212} software library.
For convolutional layers, PULP-TrainLib running on the GAP9 achieves a peak performance efficiency of 5.3 and 4.6 multiply-accumulate operations (MAC) per clock cycle for, respectively, the forward and backward passes.

\subsection{Baseline models}
\begin{figure*}[t]
  \centering
  \includegraphics[width=\textwidth]{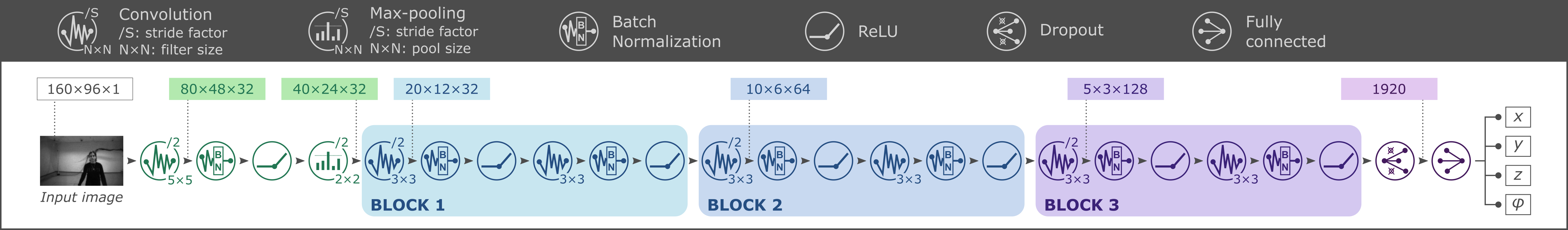}
  \caption{PULP-Frontnet~\cite{pulp-frontnet}, our target CNN architecture with 9 layers and \SI{304}{\kilo\nothing} parameters. Inference requires \SI{14.3}{\mega\mac} operations per frame.}
  \label{fig:pulp_frontnet}
\end{figure*}

We demonstrate our approach on the PULP-Frontnet CNN~\cite{pulp-frontnet}, a field-proven model for human pose estimation aboard nano-UAVs.
The architecture, in Figure~\ref{fig:pulp_frontnet} is composed of eight convolutional layers, based on the pattern \textit{conv, batch norm, ReLU}, and one fully connected,  for a total of \SI{304}{\kilo\nothing} parameters.
The total computational load for inference is \SI{14.3}{\mega\mac} (million \si{\mac}) per frame, which leads to a throughput of \SI{48}{\hertz} while consuming only \SI{96}{\milli\watt} when deployed with 8-bit integer quantization on the GAP8.
By comparison, computing one update step (forward + backward) for the entire architecture requires $3.7\times$ as many operations (\SI{53.1}{\mega\mac}).

\begin{figure}[t]
  \centering
  \rebuttalfigure{\columnwidth}{background/pulp_architecture}
  \caption{The parallel ultra-low power System-on-Chip architecture (PULP). Cluster cores perform parallel computationally intensive workloads, while a fabric controller orchestrates data transfers through two direct memory access (DMA) units. Optional floating-point units (FPU) are shared in the cluster.}
  \label{fig:pulp_architecture}
\end{figure}

\textbf{Initial training:}
we employ two datasets for human pose estimation in the initial training, from which we obtain two baseline models.
The first~\cite{pulp-frontnet} is acquired in the real world and contains \SI{2.6}{\kilo\nothing} training samples with \SI{}{\milli\meter}-precise mocap-based labels.
The second~\cite{crupi2023simtoreal} provides \SI{75}{\kilo\nothing} images and ground-truth labels from the Webots simulator.
In line with the state of the art~\cite{pulp-frontnet}, we train the models with the Adam optimizer at a learning rate of $10^{-3}$ for 100 epochs, choosing the best model during validation.
\begin{table*}[t]
    \centering
    \caption{Memory and computational requirements of fine-tuning methods}
    \label{tab:finetune_workload}
        \renewcommand{\arraystretch}{1.25}
        \begin{tabular}{cc@{\hspace{0.5em}}cc@{\hspace{0.5em}}cc@{\hspace{0.5em}}cc@{\hspace{0.5em}}cccccc}
        \toprule
        \multirow{2}[3]{*}{\shortstack[c]{\textbf{Fine-tuning}\\\textbf{strategy}}} &
        \multicolumn{2}{c}{\multirow{2}[3]{*}{\shortstack[c]{\textbf{Params}\\\textbf{[\si{\kilo\nothing}]}}}} &
        \multicolumn{7}{c}{\textbf{Memory [\si{\kilo\byte/\frame}]}} & 
        \multicolumn{4}{c}{\textbf{Computation [\si{\mega\mac/\frame/step}]}} \\
        \cmidrule(lr){4-10}\cmidrule(lr){11-14}
        & & &
        \multicolumn{2}{c}{Input} & \multicolumn{2}{c}{Activations} & \multicolumn{2}{c}{Weight grads} & \textbf{Tot.} & 
        FW & BW-IG & BW-WG & \textbf{Tot.} \\
        \midrule
        \trainall      &         304.4 & (100\%) &          15.0 & (u8) &         870.0 & (f32) &        1188.9 & (f32) &         2073.9 &            14.3 &   24.5 &   14.3 &           53.1 \\
        \trainbn       &           1.0 & (0.33\%) &         15.0 & (u8) &         585.0 & (f32) &           3.8 & (f32) &          603.8 &            14.3 &   24.4 &   0.10 &           38.8 \\
        \trainbias     & \bfseries 0.5 & (0.15\%) &         15.0 & (u8) & \bfseries 0.8 &  (u1) & \bfseries 1.9 & (f32) & \bfseries 17.7 &            14.3 &   24.4 &     -- &           38.7 \\
        \trainfc       &           7.7 & (2.5\%) & \bfseries 1.9 & (u8) &\multicolumn{2}{c}{--} &          30.0 & (f32) &           31.9 & \bfseries  0.01 &     -- &   0.01 & \bfseries 0.02 \\
        \bottomrule
        \end{tabular}
\end{table*}

\section{Method}
\label{sec:method}

\subsection{On-device fine-tuning}
\label{sec:method_ondev}
For our fine-tuning procedure, we implement the full back-propagation algorithm on an ultra-low power embedded device.
As in~\cite{NADALINI2023212}, we consider a feed-forward neural network of $N$ layers.
Each layer computes a non-linear function $f$ parameterized by a weight tensor $W_i$:
\begin{equation}
    Y_i = f_{W_i}(X_i) \qquad\qquad i \in {0, \dots, N-1}
\end{equation}
where $X_i$ and $Y_i$ are input and output tensors.
The network is a composition of the layers, where each layer's input $X_i = Y_{i-1}$ is the previous layer's output, up to the model's input $X_0$.

We train the network on a loss function $\mathcal{L}$ defined on the model outputs.
An optimization procedure, \eg stochastic gradient descent (SGD), iteratively updates the weights $W_i$ to minimize $\mathcal{L}$ according to the update rule:
\begin{equation}
\label{eq:sgd}
    W_i \leftarrow W_i + \eta \mathit{dW}_i
\end{equation}
based on the gradients of the loss function w.r.t. each weight, $\mathit{dW}_i = \delta\mathcal{L}/\delta{W_i}$, and a learning rate $\eta$.

Back-propagation enables efficient computation of the gradients through the gradient chain rule.
It performs a forward pass (FW), which computes the model's output $Y_{N-1}$ and the loss function, followed by a backward pass (BW).
For each layer starting from the last, the BW pass computes the input gradient $\mathit{dX}_i$, used to propagate the errors to the previous layers (BW-IG), and the weight gradient $\mathit{dW}_i$, used to update the layer's own parameters (BW-WG).

The trainable parameters for affine layers, such as convolution, fully connected, and batch normalization, are weight $W_i$ and bias $B_i$ tensors. Thus, their BW-IG and BW-WG steps correspond, respectively, to the products:
\begin{align}
    \mathit{dX}_i &= \mathit{dY}_i \cdot \delta{Y_i}/\delta{X_i} = \mathit{dY}_i \cdot W_i^T \\
    \rule{0pt}{3ex}
    \mathit{dW}_i &= \delta{Y_i}/\delta{W_i} \cdot \mathit{dY}_i = X_i \cdot \mathit{dY}_i \\
    \mathit{dB}_i &= \delta{Y_i}/\delta{B_i} \cdot \mathit{dY}_i = 1 \cdot \mathit{dY}_i
\end{align}
Crucially, BW-IG and BW-WG make neural network training $\sim$3$\times$ as computationally expensive as inference (which executes only the FW phase).
In addition, computing $\mathit{dW}_i$ requires storing intermediate outputs $X_i$ from the FW to the BW pass, quickly making training unfeasible on memory-constrained embedded devices.
Finally, weights and weight gradients typically have order-of-magnitude differences in scale, which complicates resorting to reduced precision arithmetic (16-bit float or quantized int).
Thus, we focus our implementation to 32-bit float arithmetic.

\textbf{Fine-tuning strategies:}
\rebuttal{
Our hardware limitations motivate us to investigate strategies to reduce the cost of the fine-tuning workload.
We select four fine-tuning strategies from the family of sparse updates, adhering to two design constraints:
\textit{i)} each strategy updates a fixed subset of the model parameters, 
\textit{ii)} each subset contains a uniform type of parameters (\eg batch normalization, fully connected, or biases).
Under these constraints, the strategies follow from optimizing different regression performance/memory/computation trade-offs due to the characteristics of each layer’s forward and backward phases.
In Table~\ref{tab:finetune_workload}, we analyze the workload of one optimization step (FW + BW) with our architecture on a single frame, depending on which subset of model parameters is updated.}

\textit{(a) Entire model (\trainall{})}:
our baseline setting is to update the weight and biases of every layer in the model, the most expensive method at \SI{53}{\mega\mac} per frame.
In addition, it requires storing all intermediate activations and weight gradients, which in 32-bit floats amount to \SI{2}{\mega\byte} for a single frame, 25\% of the AI-deck's \SI{8}{\mega\byte} DRAM, which is also needed to store our fine-tuning dataset.

\textit{(b) Only batch-norm layers (\trainbn{})}:
batch normalization layers contain a mere 0.33\% parameters of the entire model while significantly impacting model performance~\cite{mudrakarta2018k}.
This method requires 70\% less memory and 26\% less computation due to the smaller weight gradients.

\textit{(c) Only biases (\trainbias{})}:
as BW-WG of the biases is simply the BW-IG of the following layer, full intermediate activations don't need to be stored~\cite{cai2020tinytl}.
The 1-bit signs of the intermediate activations are still needed for BW-IG of ReLU non-linearities, \SI{0.8}{\kilo\byte} per frame.
Total memory decreases to just \SI{17.7}{\kilo\byte}, while computation remains comparable to \trainbn{}.

\textit{(d) Only the fully-connected layer (\trainfc{})}:
by far, the greatest reduction comes from fine-tuning only the final layer of the network, although this also has the least expected improvement among the four methods~\cite{cai2020tinytl}.
The forward pass of all layers up to the final fully connected one can be pre-computed just once at the beginning of fine-tuning, a 99.9\% reduction in computation.
In our implementation, this pre-computed forward pass is further quantized to 8-bit integers and runs in real-time at \SI{48}{\hertz} during fine-tuning set acquisition.
The fine-tuning dataset then only stores the resulting feature vectors (1920 elements) instead of full input images ($160 \times \SI{96}{\pixel}$), reducing memory storage by 98.1\%.

\textbf{Hyper-parameters:}
compared to the initial training, we substitute Adam with stochastic gradient descent (SGD) due to $3\times$ lower computation cost. With hyper-parameter search, we set the learning rate to $10^{-2}$ to retain comparable regression performances to Adam.
Additionally, we select a fine-tuning set size of 512 samples, such that the entire dataset fits the AI-deck's \SI{8}{\mega\byte} DRAM memory even for methods that require storing the full input images (\SI{7.8}{\mega\byte}), and set five epochs as fixed fine-tuning length.
\rebuttal{We always select the final model at the end of fine-tuning, thus avoiding storing multiple model checkpoints.}

\begin{figure*}[t]
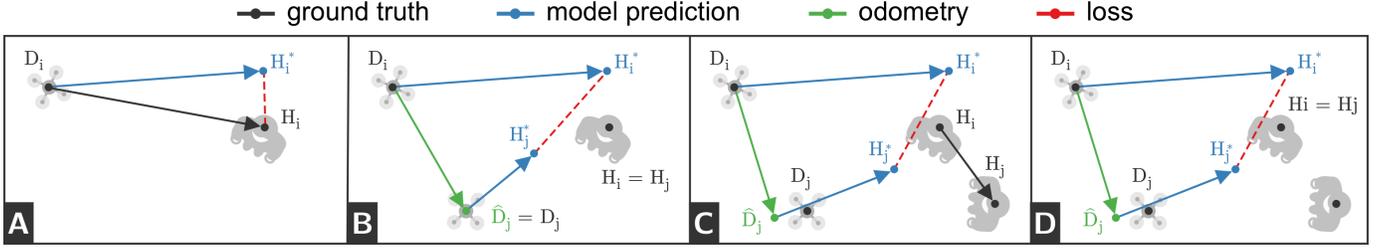

  \centering
  \rebuttalfigure{\textwidth}{method/ref_frames}
  \caption{Loss functions: A) task loss, B) original state-consistency loss~\cite{nava2021uncertainty}, and C-D) our state-consistency loss with uncertain drone odometry and moving subject. Subject movements are either C) known or D) unknown (subject assumed still). The depicted reference frames represent drone $\mathrm{D}$ and subject $\mathrm{H}$ ground-truth poses, at times $i$ and $j$, and the corresponding relative poses estimated by, respectively, drone odometry $\mathrm{\hat{D}}$ and model predictions $\mathrm{H^*}$.}
  \label{fig:ref_frames}
\end{figure*}

\textbf{Data augmentation:}
we employ standard photometric data augmentations, including exposure and contrast adjustment, Gaussian noise, box blurring, and vignetting.
These transformations operate mostly pixel-wise, benefiting from parallelization and high spatial locality, allowing efficient implementations on embedded devices such as GAP SoCs.
Furthermore, we randomly flip images and ground-truth labels horizontally to guarantee the fine-tuning set follows a symmetric distribution along the $y$ axis and yaw orientation.
Similarly, we randomly apply time reversal when fine-tuning with state consistency to impose that relative poses $\relpose{D_i}{D_j}$ and $\relpose{H_i}{H_j}$ also respect a zero-mean symmetric distribution (i.e., centered on the identity).

\subsection{Self-supervised loss}
\label{sec:method_ssl}

We perform self-supervised learning with the state-consistency loss introduced by Nava \textit{et al.}~\cite{nava2021uncertainty}.
\rebuttal{We consider a dataset of fine-tuning samples acquired at timesteps $i \in \mathcal{T}$, where each sample comprises an image from the drone's camera and the drone's and subject's poses.}
We define $\relpose{A}{B}$ in $\mathrm{SE}(3)$ as the relative pose of reference frame $B$ w.r.t. $A$.
The fine-tuning process minimizes the loss function
\begin{equation}
\label{eq:loss}
    \mathcal{L} = \mathcal{L}_\mathrm{task} + \lambda_\mathrm{sc} \mathcal{L}_\mathrm{sc},
\end{equation}
composed of a task loss term $\mathcal{L}_\mathrm{task}$ and a state-consistency loss term $\mathcal{L}_\mathrm{sc}$.
In accordance with~\cite{nava2021uncertainty}, we set $\lambda_\mathrm{sc} = 1$.

\rebuttal{Figure~\ref{fig:ref_frames} depicts the two loss terms and their effects on model predictions.}
The \textbf{task loss} is defined on individual timesteps $i$, taken from the (possibly empty) subset $\mathcal{T}_\mathrm{t} \subseteq \mathcal{T}$ for which target relative poses $\relpose{D_i}{H_i}$ are known:
\begin{equation}
\label{eq:task_loss}
    \mathcal{L}_\mathrm{task} = \frac{1}{|\mathcal{T_\mathrm{t}}|} \sum_{i \in \mathcal{T}_t} \Delta(\relpose{D_i}{H^*_i}, \relpose{D_i}{H_i}),
\end{equation}
where $\relpose{D_i}{H^*_i}$ represents the model's estimate of the subject w.r.t. the drone relative pose at time $i$ and $\Delta(\mathbf{T_1}, \mathbf{T_2})$ is a distance function between relative poses, which we define as the L1 loss between 4DOF pose vectors $(x, y, z, \phi)$\footnote{Angles are expressed in radians and angle differences are computed on the circle group, to account for discontinuities at $\pm\pi$. This weighs equally position errors of 1 m and rotation errors of \SI{1}{\rad}, which is a reasonable heuristic in our setting.}.

When $\relpose{D_i}{H_i}$ are ground-truth relative poses (e.g., acquired by a motion capture system), this loss is equivalent to ordinary supervised learning. Noisy estimates can also be used as $\relpose{D_i}{H_i}$.
Our experiments explore the case when the relative pose is known at a time $i$, and the subject subsequently remains still, which allows us to define the relative pose at a later time $j$ as $\relpose{\hat{D}_j}{H_j} = \relpose{\hat{D}_j}{D_i} \relpose{D_i}{H_i}$. The relative pose $\relpose{D_i}{\hat{D}_j}$ indicates the (possibly noisy) odometry estimate of the drone's pose at time $j$ w.r.t time $i$.

The \textbf{state-consistency loss} is defined instead on pairs of timesteps $i$ and $j = i + \mathit{dt}$ at a fixed time delta (a hyperparameter) sampled from the subset $\mathcal{T}_\mathrm{sc} \subseteq \mathcal{T}$:
\begin{equation}
\label{eq:sc_loss}
    \mathcal{L_\mathrm{sc}} = \frac{1}{|\mathcal{T_\mathrm{sc}}|}\sum_{i \in \mathcal{T_\mathrm{sc}}} \Delta(\relpose{H^*_i}{D_i} \relpose{D_i}{\hat{D}_j} \relpose{D_j}{H^*_j}, \; \relpose{H_j}{H_i}),
\end{equation}
where $\relpose{H_j}{H_i}$ is the subject's relative pose at time~$j$ w.r.t time~$i$.

\rebuttal{Compared to \cite{nava2021uncertainty}, we reformulate the state-consistency loss to model drone and subject movements separately.
Noisy estimates for the former are generally available through drone odometry, while the latter are not known by the drone.
In the experiments, we evaluate the regression performance impact of these sources of uncertainty individually as depicted in Figure~\ref{fig:ref_frames}-BCD. 
Subject movements $\relpose{H_j}{H_i}$ will be replaced by the identity matrix $\ident$, \textit{i.e.}, we assume the subject stands still.}

\begin{figure*}[t]
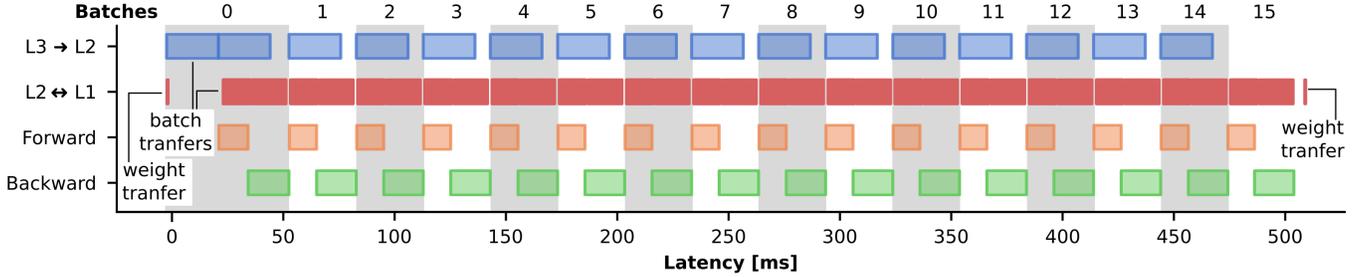

\centering
\rebuttalfigure{\textwidth}{method/pipeline}
\caption{\rebuttal{One epoch of the fine-tuning pipeline on GAP9 with all optimizations (8 cores, L3-L2 double buffering, loop unrolling, and L1 memory).}}
\label{fig:pipeline}
\end{figure*}

\subsection{Embedded implementation}
\label{sec:method_impl}
\rebuttal{We implement the proposed fine-tuning process in optimized C code for the PULP architecture.
Figure~\ref{fig:pipeline} shows the algorithm pipeline, depicting dependencies and concurrent operations.
We compute the forward, loss function, and backward passes using hand-written 32-bit float kernels that operate entirely in L1 memory.
The 512-sample fine-tuning set, composed of input images and the corresponding 4-element label outputs, totals \SI{7.8}{\mega\byte} stored as 8-bit quantized integers in the off-chip L3 memory.
The \trainfc{} method, which only ever needs to perform the forward pass before the fully-connected layer once (see Section~\ref{sec:method_ondev}), stores just the fully-connected layer's 1920-element input features and label outputs ($\sim$\SI{1}{\mega\byte}).}

\rebuttal{During fine-tuning, the dataset is processed in 16 batches of 32 samples each.
Each batch is transferred from L3 to L2 memory, while samples are transferred to L1 memory one by one.
Both transfers exploit the SoCs' DMA peripherals.
Weight gradients are accumulated in L1 memory and applied to the weights once per batch.
The updated weights are written out to L2 at the end of each epoch.}

\rebuttal{
\textbf{Optimizations:}
In our kernel implementations, we apply four software optimizations to exploit the target platform's hardware architecture efficiently.\\
\textit{i) Double buffering:}
L3-L2 memory transfers are constrained by the limited off-chip memory bandwidth ($\sim$\SI{90}{\mega\byte/\second}).
As shown in Figure~\ref{fig:pipeline}, we introduce double buffering to schedule DMA transfers concurrently with forward/backward computation, thereby fully hiding transfer latency (except for the first batch in each epoch).\\
\textit{ii) Loop unrolling:}
on GAP chips (strictly in-order architectures), memory instructions suffer from load stalls due to access latency. To hide these latencies, we manually unroll 4 iterations of our kernels along the output tensor dimension.\\
\textit{iii) L1 memory:}
We take advantage of the cluster L1 memory area (\SI{64}{\kilo\byte} on GAP8, \SI{128}{\kilo\byte} on GAP9, single-cycle access time) for every buffer (\ie input, outputs, labels, weights, and weight gradients) in the inner loops of our kernels.
The cluster DMA handles L2-L1 transfers sample-by-sample.\\
\textit{iv) Parallelization:}
We take advantage of the GAP chips' multi-core cluster by executing the forward/backward computation on 8 cores, with a blocking strategy that distributes work among cores along the input tensor dimension.
Section~\ref{sec:results_ondev} measures the latency improvements of each optimization individually.}

\newlength{\oldcolumnwidth}
\setlength{\oldcolumnwidth}{\columnwidth}
\setlength{\columnwidth}{0.74\columnwidth}
\begin{table}[t]
    \centering
    \rebuttaltable
    \caption{Average regression performance}
    \label{tab:regression_metrics}
    \resizebox{\columnwidth}{!}{
      \renewcommand{\arraystretch}{1}
      \begin{tabular}{ccccccl}
          \toprule
          \multirow{1}{*}{\textbf{Train on}} &
          \multirow{1}{*}{\textbf{Fine-tune on}} & 
          \multirow{1}{*}{\textbf{MAE}} & 
          \multirow{1}{*}{$\mathbf{R^2}$ \textbf{[\%]}} &
          \multirow{1}{*}{\textbf{cfr. Fig.~\ref{fig:r2_matrix}}}
          \\
          \midrule
          \multirow{3}{*}{\shortstack[c]{real\\world}} & --       & 0.50 &  -9.3 & (A), SotA~\cite{pulp-frontnet} \\
                                                       & env.   & 0.39 &  23.5 &  -- \\
                                                       & env. + subj.       & 0.35 &  41.5 &  -- \\
          \midrule
          \multirow{3}{*}{sim.} & --       & 0.61 & -55.0 & (B), SotA~\cite{crupi2023simtoreal} \\
                                & env.   & 0.38 &  29.1 & (C) \\
                                & env. + subj.       & \bfseries 0.27 & \bfseries 57.4 & (D) \\
          \bottomrule
      \end{tabular}
    }
\end{table}
\setlength{\columnwidth}{\oldcolumnwidth}

\FloatBarrier % Force table and figure on two different columns

\begin{figure}[t]
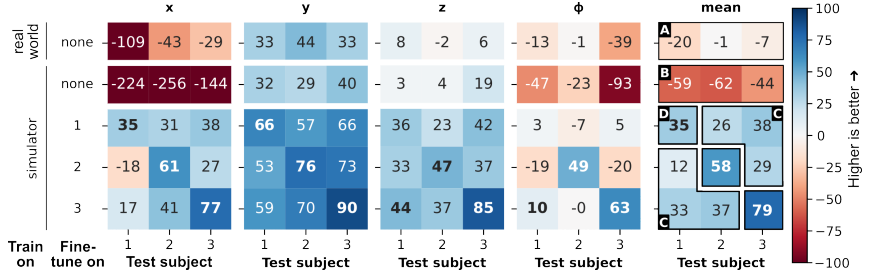

  \rebuttalfigure[right]{1.28\columnwidth}{results/r2_matrix}
  \caption{$R^2$ scores [\%] for each combination of fine-tuning and test subjects.}
  \label{fig:r2_matrix}
\end{figure}

\rebuttal{
\textbf{Quantization:}
At inference time, we run the int-8 quantized model obtained through Parameterized Clipping Activations (PACT)~\cite{choi2018pact} and Quantization-Aware Training (QAT).
The quantization process is first performed offline on the non-finetuned baselines through the same deployment pipeline adopted by the original works~\cite{pulp-frontnet,crupi2023simtoreal}.
These quantized initial model parameters are deployed to our SoC's Flash memory.
At fine-tuning time, the quantized weights are loaded and dequantized to a 32-bit floating-point in the on-chip L2 memory.
Fine-tuning is then performed entirely in floating-point arithmetic.
Updated weights are quantized again at the end of the fine-tuning process to the same 8-bit integer representation used for inference.
In particular, we found the overall scale of model parameters does not change significantly during fine-tuning. As such, we quantize using the same quantization parameters (quantization $\varepsilon$ and bit precision) as the non-finetuned model, without repeating QAT.}
\section{Results} \label{sec:results}

\subsection{Fine-tuning strategy} \label{sec:offline-exp}
We start our experiments by analyzing the proposed fine-tuning methods offline in the PyTorch framework on a set of \SI{4.7}{\kilo\nothing} real-world images from~\cite{cereda2022pitchaug}.
This dataset comprises 18 in-flight sequences from three subjects in challenging situations, e.g., different subject appearances, dynamic motions of both subjects and the drone.
It is collected with the same robotic platform introduced in Section~\ref{sec:background} (including the Himax camera) and is used both for fine-tuning the pre-trained models and testing them.
This dataset stresses the domain shift problem, as it represents a completely novel domain for the baseline models, which are trained either entirely in simulation or in a different real-world environment.

For each subject, we select a random temporally-contiguous \SI{128}{\second} segment of the dataset (512 samples @ \SI{4}{\hertz}) as the fine-tuning set, while the rest is used as the test set.
To provide unbiased measures of regression performance, we discard 100 contiguous samples (\SI{25}{\second}) between fine-tuning and test segments, and we apply cross-validation, repeating 3 runs for each subject with different random fine-tuning segment (9 total experiment runs).
At most, 75\% of each subject's samples are used as the fine-tuning set.

\textbf{Baseline performance:}
A lower bound on the test regression performance achieved by our models is obtained by not doing any fine-tuning. 
An upper bound corresponds to fine-tuning in the best-case scenario: we fine-tune all model parameters, i.e. method \trainall, entirely on the task loss (\textit{i.e.}, $\mathcal{T}_\textrm{t} = \mathcal{T}$ and $\mathcal{T}_\textrm{sc} = \varnothing$) and assuming perfect knowledge of the drone and subject poses.

Table~\ref{tab:regression_metrics} reports the resulting mean absolute errors (MAE) and $R^2$ scores\footnote{The $R^2$ score is a normalized metric of regression performance that aids comparisons among different sets of data. A perfect model scores 100\%, while a dummy model always predicts the test set mean scores zero.}. \rebuttal{Figure~\ref{fig:r2_matrix} breaks down these results} by the test subject, both on the four individual regression outputs and as an average.
Model (\textit{A}), trained on real-world data and not fine-tuned, represents the performance of the original PULP-Frontnet~\cite{pulp-frontnet} on our test set.
Model (\textit{B}) is trained on simulated data, is not fine-tuned, and constitutes our lower bound.

For the fine-tuned models, we report several averages: (\textit{D}) reports the average performance of a model when tested on the same subject and the same environment that it was fine-tuned on; (\textit{C}) is the average performance of a model, tested on a different subject (but the same environment) that it was fine-tuned on.
Table~\ref{tab:regression_metrics} shows that, despite the higher initial performance, the models trained with real-world data achieve a poorer fine-tuned performance than those trained on simulation data.
This confirms that training on a larger and more varied simulation train set translates to a model that better adapts to unseen subjects and environments when fine-tuned.
Therefore, in the following experiments, we focus on scenario (\textit{D}): models trained on simulated datasets and tested on the same subject and environment they were fine-tuned on.

\begin{figure}[t]
  \centering
  \includegraphics[width=\columnwidth]{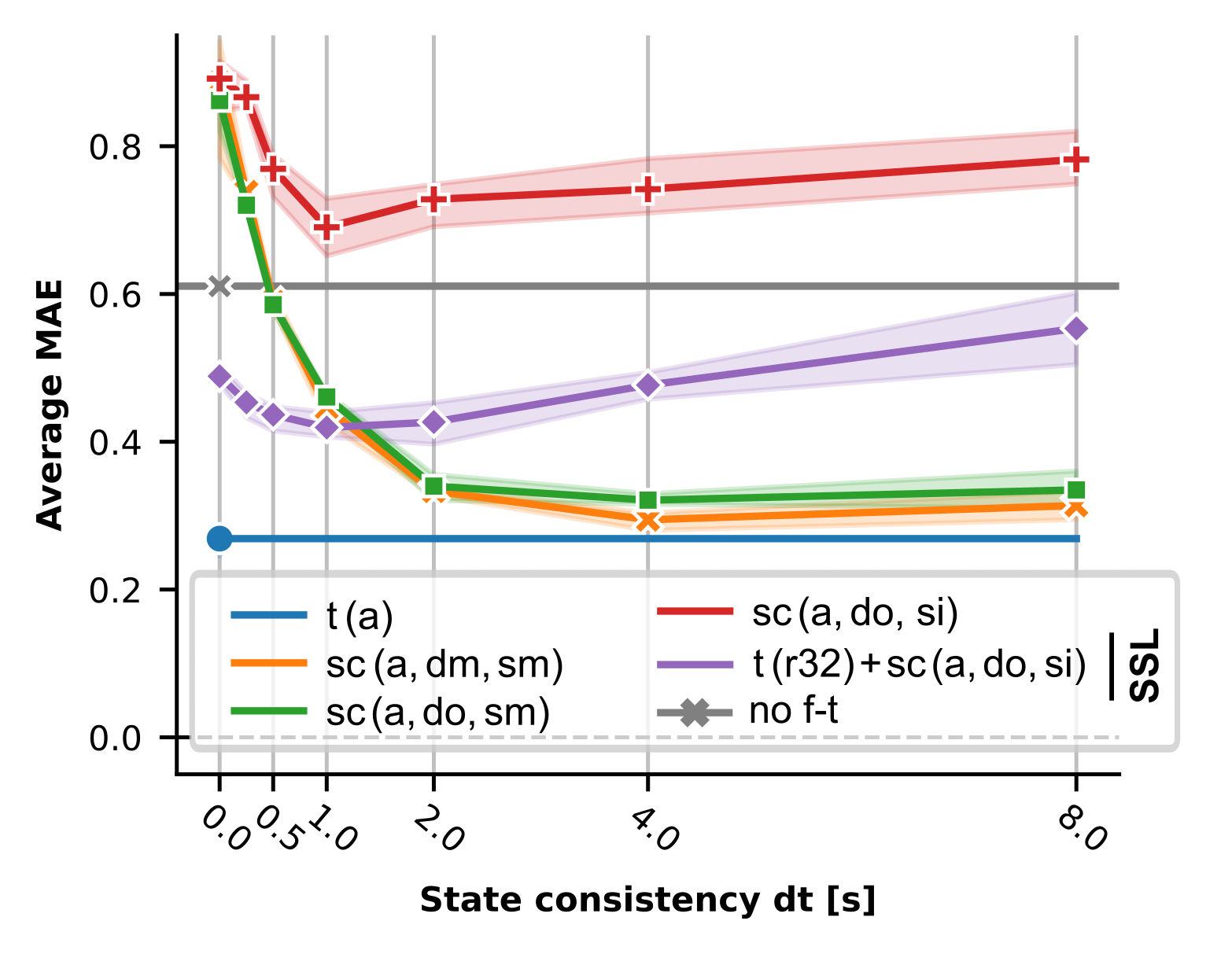}
  \caption{Regression performance across loss functions. The self-supervised loss (SSL) is computed entirely on data from the drone's onboard sensors and reaches up to 50\% of the MAE improvement of the supervised loss $t(a)$.}
  \label{fig:loss}
\end{figure}

\textbf{Self-supervised learning:}
in Figure~\ref{fig:loss}, we report our fine-tuning results obtained when ground-truth labels are unavailable.
We consider three setups for the drone pose -- \textit{perfect absolute pose}, \textit{perfect odometry} (dm), and \textit{uncertain odometry} (do) -- and three for the pose of the human subject -- \textit{perfect absolute pose}, \textit{perfect odometry} (sm), and \textit{unknown} (si).

When perfect absolute poses are known for both drone and subject, we have the ground-truth information to fine-tune using $\mathcal{L}_\mathrm{t}$ (\textit{i.e.}, regular supervised learning).
The \textit{\MakeLowercase{\lossid{}}} case where these are known for all samples, considered in all previous experiments, is named $\mathrm{t(a)}$ and reaches MAE 0.27.

To reduce our reliance on privileged information, we assume only odometry is known, \textit{i.e.} relative poses between two instants in time.
When odometry is perfect for both drone and subject on all samples $\mathrm{sc(a, dm, sm)}$, we can fine-tune using $\mathcal{L}_\mathrm{sc}$, which achieves 93\% of the ideal improvement.
Uncertain drone odometry $\mathrm{sc(a, do, sm)}$ also has a limited impact on performance and achieves 85\% of the ideal.
Performance improves at higher state-consistency time deltas $\mathit{dt}$ (horizontal axis), as imposing state-consistency between samples farther in time carries a larger information content. 

On the other hand, an unknown subject odometry $\mathrm{sc(a, do, si)}$ drastically reduces performance\footnote{The case with unknown subject odometry but perfect drone odometry $\mathrm{sc(a, dm, si)}$, omitted, exhibits similar behavior.}.
In this case, in order to compute $\mathcal{L}_\mathrm{sc}$, we assume that $\relpose{H_i}{H_j} = \ident$, \textit{i.e.}, that the subject is always still.
Although the time reversal augmentation ensures this holds on average, \textit{i.e.}, the fine-tuning set has $\mathbb{E}[\relpose{H_i}{H_j}] = \ident$ by design, the model degenerates to a dummy predictor that always outputs a constant value.

To address the above issue, we design a cooperative scenario in which the subject moves to a known pose w.r.t. the drone (\eg \SI{1}{\meter} in front of the drone, directly facing the camera and centered in the field of view).
While the subject stands still, the drone randomly moves around to acquire a small number of samples.
The procedure is repeated from multiple start locations in the environment to acquire highly diverse fine-tuning data.
We test this scenario, called $\mathrm{t(r32) + sc(a, do, si)}$, by selecting a random 32-sample subset of frames on which we optimize $\mathcal{L}_t$.
It relies on realistic in-field infrastructure-free data acquisition and achieves a significant improvement, up to 39\%. % of the ideal case.
As such, we select it as the \lossre{} loss function in the following experiment, in the best-performing $\mathit{dt} = \SI{2}{\second}$ configuration.

\begin{table}[t]
    \centering
    \caption{Regression performance compared across fine-tuning methods}
    \label{tab:finetune_method}
    \resizebox{\columnwidth}{!}{
        \renewcommand{\arraystretch}{1.3}
        \begin{tabular}{ccccccc}
        \toprule
         & \textbf{Fine-tuning} & none & \trainall & \trainbn & \trainbias & \trainfc \\
        \midrule
        \parbox[t]{2mm}{\multirow{2}{*}{\rotatebox[origin=c]{90}{\textbf{MAE}}}}
        & \lossid & 0.61 & \bfseries 0.27 &           0.36 & 0.39 & 0.45 \\
        & \lossre & --   & \bfseries 0.43 & \bfseries 0.43 & 0.46 & 0.47 \\
        \bottomrule
        \end{tabular}
    }
\end{table}

\textbf{Fine-tuning methods:} in Table~\ref{tab:finetune_method}, we explore the effectiveness of methods that reduce the fine-tuning workload by limiting the subset of model parameters to update.
Full fine-tuning, named \trainall, sets the lower bound at \rebuttal{an MAE} of 0.27 (\SI{-56}{\percent} compared to the non-finetuned baseline).
Optimizing only the batch-norm layers, \trainbn, is second best, followed closely by fine-tuning the biases, \trainbias.
Fine-tuning the final fully connected layer, \trainfc, performs the worst, but notably still shows \rebuttal{an MAE} improvement of \SI{-26}{\percent}.

\subsection{In-field experiments}
\label{sec:infield}
In this experimental section, we assess the end-to-end performance of our system when deployed in the real world, i.e., in the field.
We challenge our models in an environment and test subject combination that significantly differs from those seen during initial training.
As described above, we collect a 512-sample fine-tuning dataset, then fine-tune eight models: four fine-tuning methods, \trainall{}, \trainbn{}, \trainbias{}, and \trainfc{}, multiplied by two loss functions, supervised and self-supervised.
We record three flights for each of our two baselines and eight fine-tuned models ($3 \times 10 = 30$ flights), plus one additional flight based on the perfect mocap position of the subject, for a total of 31 test flights.
As the GAP9 SoC is not currently available on a COTS module for the Crazyflie, for in-field experiments, we deploy all fine-tuned models on GAP8, applying 8-bit integer quantization to achieve a real-time inference throughput of \SI{48}{frame/\second}.

\begin{table}[t]
    \centering
    \caption{In-field experiment results (average over 3 runs).\\The symbol * marks prematurely failed runs.}
    \label{tab:infield_results}
    \resizebox{\columnwidth}{!}{
      \renewcommand{\arraystretch}{1.25}
      \begin{tabular}{clcccccc} % <-- Alignments: 1st column left, 2nd middle and 3rd right, with vertical lines in between
        \toprule
        & \multirow{2}[3]{*}{\textbf{Model}} & \multirow{2}[3]{*}{\shortstack[c]{\textbf{Completed}\\ \textbf{path [\%]}}} & \multicolumn{3}{c}{$\mathbf{R^2}$ \textbf{[\%]}} & \multicolumn{2}{c}{\textbf{Control error}} \\
        \cmidrule(lr){4-6} \cmidrule(lr){7-8}
        & & & $x$ & $y$ & ${\phi}$  & $e_{xy}$ [m] & $e_\theta$ [rad] \\
        \midrule
        & {mocap} & 100 & 100.0 & 100.0 & 100.0 & 0.18 & 0.21 \\
        \midrule
        \parbox[t]{2mm}{\multirow{2}{*}{\rotatebox[origin=c]{90}{\rebuttal{\textbf{SotA}}}}}%
        & {real world}~\cite{pulp-frontnet}      & 0 & -80.7 & 26.3 & 1.9 & 2.68* & 0.32* \\
        & {simulator}~\cite{crupi2023simtoreal}  & 91 & 9.0 & 42.1 & 12.7 & 1.05* & 0.37* \\
        \midrule
        \parbox[t]{2mm}{\multirow{4}{*}{\rotatebox[origin=c]{90}{\textbf{\lossid}}}}%
        & \trainall  & 100 & \textbf{64.5} & \textbf{75.8} & \textbf{24.6} & 0.70 & 0.47 \\
        & \trainbn   &  76 & 53.4 & 62.3 & 20.5 & 0.60* & 0.30* \\
        & \trainbias & 100 & 46.3 & 55.2 & 18.8 & 0.72 & \textbf{0.46} \\
        & \trainfc   & 100 & 44.9 & 54.5 & 15.6 & \textbf{0.67} & 0.47 \\
        \midrule
        \parbox[t]{2mm}{\multirow{4}{*}{\rotatebox[origin=c]{90}{\textbf{\lossre}}}}%
        & \trainall  &  92 & \textbf{36.2} & \textbf{55.4} & \textbf{21.1} & 1.03* & 0.45* \\
        & \trainbn   & 100 & 33.9 & 54.1 & 14.6 & 1.00 & 0.59 \\
        & \trainbias & 100 & 33.3 & 48.8 & 15.8 & 0.99 & 0.43 \\
        & \trainfc   & 100 & 32.9 & 48.0 & 15.8 & \textbf{0.78} & \textbf{0.42} \\
        \bottomrule
      \end{tabular}
    }
\end{table}

\begin{figure*}[t]
\centering
\includegraphics[width=\textwidth]{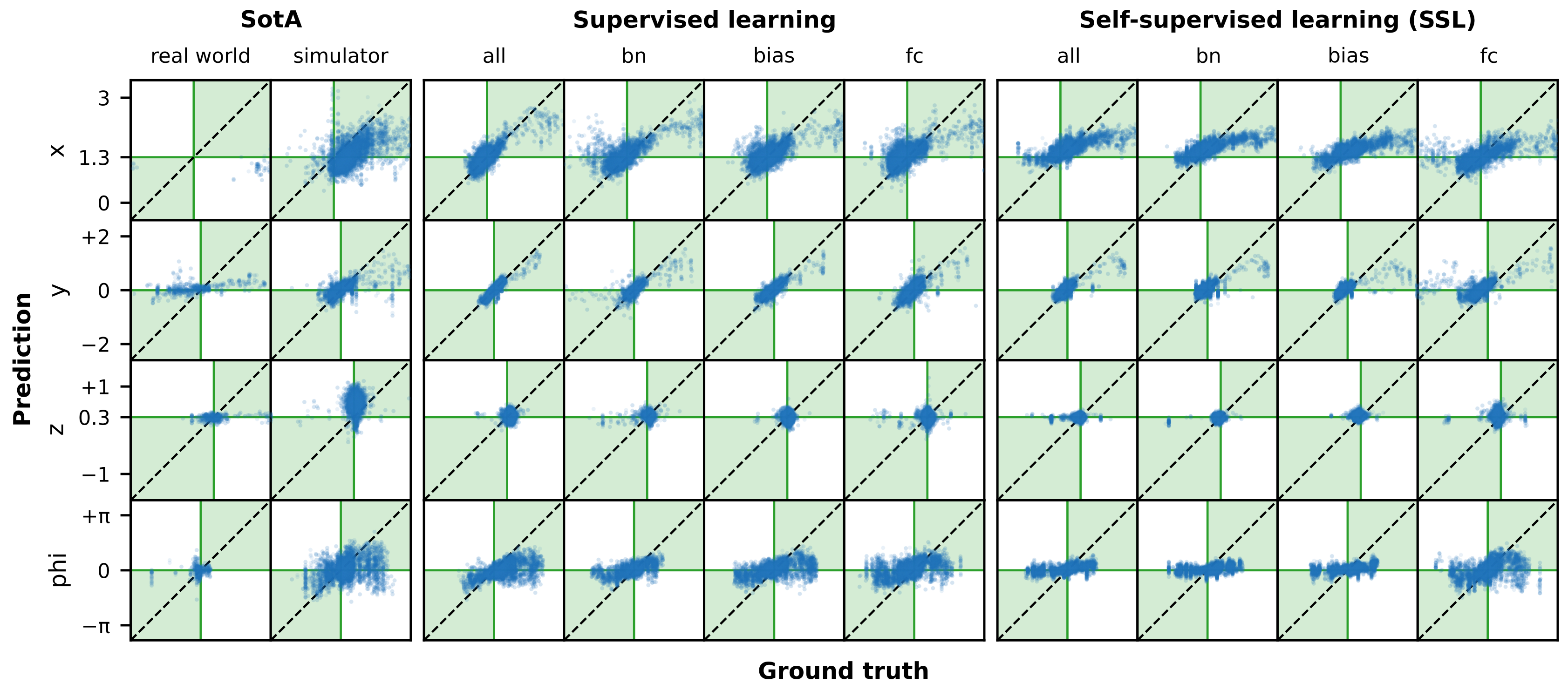}
\caption{Predictions (vertical) vs. ground-truths (horizontal) for each model during its three in-field experiment runs. Dashed diagonals correspond to a perfect predictor. Green lines mark the control setpoints: predictions in the green area contribute to drive the drone towards the desired pose.}
\label{fig:scatter-plots}
\end{figure*}

We reproduce the same testing procedure introduced in~\cite{pulp-frontnet}: the human subject walks a predetermined path, with distinct phases designed to stress the different components of the model prediction, while the fully autonomous nano-UAVs must stay in front of the subject at a distance of \SI{1.3}{\meter}.
The results are summarized in Table~\ref{tab:infield_results}, where we report the path completion (in percentage), the $R^2$ regression performance, and the control errors, all of them averaged over the three runs of each model\footnote{Supplementary video material for the in-field experiments available online: \url{https://youtu.be/3yNbMwszpSY}}.

We define path completion as the percentage of the expected path the drone can follow, and we conclude the test if the subject exits the camera's field of view or the drone leaves our $6\times6$\si{\meter} flight arena.
The SotA real-world baseline~\cite{pulp-frontnet} immediately fails the experiment: it moves away from the subject instead of approaching, thus completing $0\%$ of the path.
The SotA simulator baseline~\cite{crupi2023simtoreal} manages to follow the subject two out of three runs for the entire path, resulting in a completion score of $91\%$.
Fine-tuned models track the subject until the end of every experiment, except one run each of the supervised \trainbn{} and self-supervised \trainbias{} models.
The supervised \trainbn{} run is aborted because the drone comes close to the subject, while the drone in the self-supervised \trainall{} run drifts out of the arena.

The second metric analyzed in Table~\ref{tab:infield_results} is the $R^2$, which indicates the regression performance on a test set.
For comparability between models, we build this test set from all 31 test flights, resulting in \num{60104} images. 
The relative trends between models closely follow those in Section~\ref{sec:offline-exp}, i.e., fine-tuned models outperform the simulator baseline by up to $55\%$, and self-supervised models also manage a $27\%$ improvement.

Finally, the last metric we present in Table~\ref{tab:infield_results} is the control performance, which is composed of the mean horizontal position error $e_{xy}$ and the mean absolute angular error $e_{\theta}$.
The former is computed against the desired position of the drone (i.e., \SI{1.5}{\meter} in front of the subject) while $e_{\theta}$ accounts for the drone orientation vs. the desired one (i.e., facing the subject).
Compared to the two SotA baselines, the supervised models improve the in-field control performance by $-36\%$ on $e_{xy}$.
Self-supervised models \trainall{}, \trainbn{}, and \trainfc{} show only a minor reduction in control error compared to the SotA simulator baseline (i.e., $-6\%$ on $e_{xy}$), while \trainfc{} marks a clear improvement w.r.t the same baseline, with $e_{xy}$ of just \SI{0.78}{\meter}.

\begin{figure*}[t]
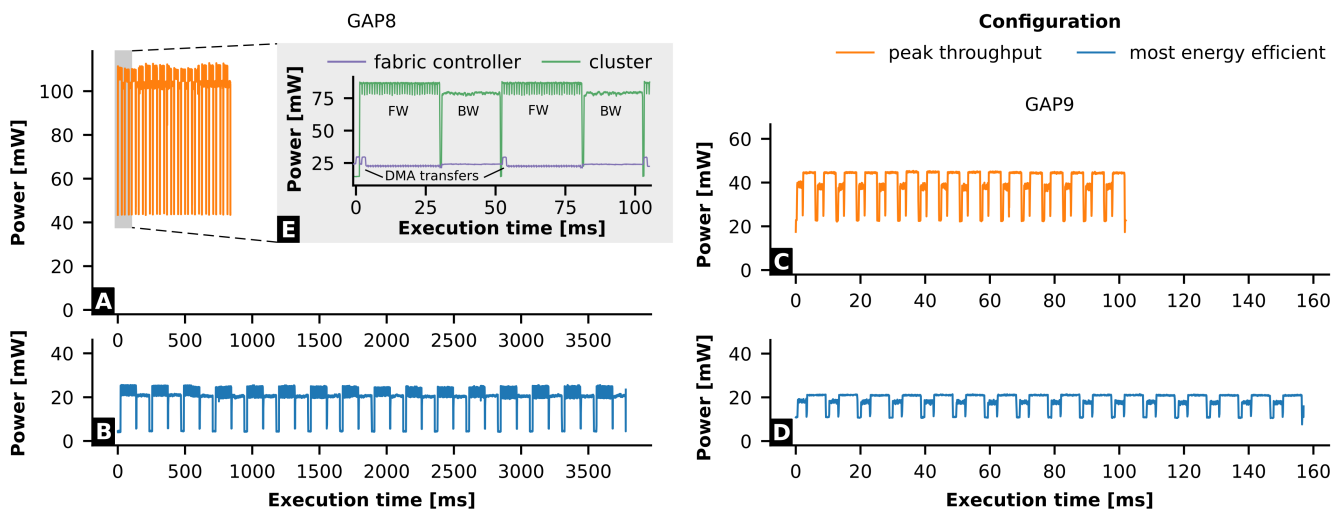

\centering
\rebuttalfigure{\textwidth}{results/power_trace_detail}
\caption{
\rebuttal{Power traces for one epoch of the \trainfc{} fine-tuning method, on A-B) GAP8 and C-D) GAP9.
SoCs are configured in the A-C) \textit{peak throughput} and B-D) \textit{most energy efficient} operating points.
E) Details of the first two batches on GAP8 at peak throughput, with power consumption broken down into fabric controller and cluster power domains.}}
\label{fig:power-trace}
\end{figure*}

This last result on the SSL \trainfc{} remarkably shows only a minor degradation (only 16\%) in control performance compared with the vastly costlier supervised \trainall, i.e., $\num{10000}\times$ more operations and 200$\times$ more memory than SSL \trainfc{}.
To further investigate this key finding, we show in Figure~\ref{fig:scatter-plots} the individual predictions of each model.
From these, we notice the real-world baseline consistently under-estimates $x$, outside the controller convergence zone (in green).
On the contrary, all fine-tuned models show improved performance compared to the baselines.
Self-supervised models reduce prediction noise even more than supervised models but show stronger miscalibration w.r.t. the ground-truths, which should be explored in future work.
The better control performance by SSL \trainfc{} w.r.t. the other three SSL models is explained by the inability of the latter three to estimate $\mathit{phi}$.

\begin{table}[t]
    \centering
    \caption{On-device fine-tuning cost (5 epochs)}
    \label{tab:ondevice-cost}
    \resizebox{\columnwidth}{!}{
        \renewcommand{\arraystretch}{1.1}
        \begin{tabular}{cccccc}
        \toprule
        \multicolumn{2}{c}{\multirow{2}{*}{\shortstack[c]{\textbf{SoC and operating point}\\(Vdd, FC and CL frequencies)}}} & 
        \multirow{2}{*}{\shortstack[c]{\textbf{Latency}\\\textbf{[\si{\milli\second}]}}} &
        \multirow{2}{*}{\shortstack[c]{\textbf{Avg. Power}\\\textbf{[\si{\milli\watt}]}}} &
        \multirow{2}{*}{\shortstack[c]{\textbf{Energy}\\\textbf{[\si{\milli\joule}]}}} \\
        \\
        \midrule
        \multirow{4}[1]{*}{\rotatebox[origin=c]{90}{\textbf{GAP8}}} &
          most energy efficient & \multirow{2}{*}{\num{18899}} & \multirow{2}{*}{24.6} & \multirow{2}{*}{464.7} \\
        & (\SI{1.0}{\volt}, \SI{25}{\mega\hertz}, \SI{75}{\mega\hertz}) & & & & \\
        \rule{0pt}{3ex}
        & peak throughput & \multirow{2}{*}{\num{6553}} & \multirow{2}{*}{101.5} & \multirow{2}{*}{664.9} \\
        & (\SI{1.2}{\volt}, \SI{250}{\mega\hertz}, \SI{175}{\mega\hertz}) & & & & \\
        \midrule
        \multirow{4}[1]{*}{\rotatebox[origin=c]{90}{\textbf{GAP9}}} &
          most energy efficient & \multirow{2}{*}{785} & \multirow{2}{*}{\textbf{18.7}} & \multirow{2}{*}{\textbf{14.7}} \\
        & (\SI{0.65}{\volt}, \SI{240}{\mega\hertz}, \SI{240}{\mega\hertz}) & & & & \\
        \rule{0pt}{3ex}
        & peak throughput & \multirow{2}{*}{\textbf{511}} & \multirow{2}{*}{38.3} & \multirow{2}{*}{19.5} \\
        & (\SI{0.80}{\volt}, \SI{370}{\mega\hertz}, \SI{370}{\mega\hertz}) & & & & \\
        \bottomrule
        \end{tabular}
    }
\end{table}

\begin{figure}[t]
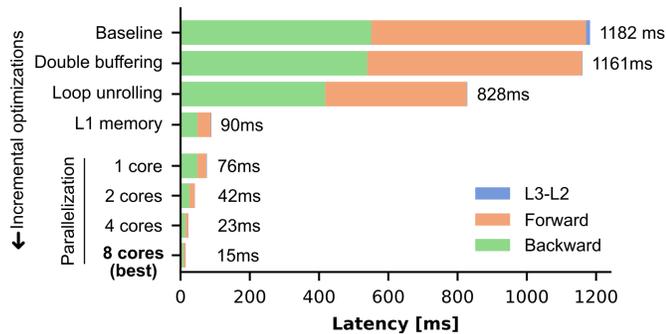

\centering
\rebuttalfigure{\columnwidth}{results/impl_optimization}
\caption{\rebuttal{Latency comparison between kernel implementations with different optimization levels. One batch on GAP9 at the ``peak performance'' operating point.}}
\label{fig:impl-optimization}
\end{figure}

\subsection{On-device fine-tuning}
\label{sec:results_ondev}
As a final experiment, we deploy the proposed method on the GAP8 and GAP9 SoCs and profile its workload.
Due to its computational and memory advantages, combined with the highest in-field self-supervised performance, we focus on \trainfc{} for on-device deployment.

Figure~\ref{fig:power-trace} shows the detailed power traces for one fine-tuning epoch at different operating points.
Instead, Table~\ref{tab:ondevice-cost} provides aggregate metrics on the fine-tuning process composed of 5 epochs.
The zoom-in on the first two batches in Figure~\ref{fig:power-trace}-C highlights the sequence of forward (FW) and backward (BW) passes with their distinct power consumption patterns.
The breakdown into fabric controller and cluster power consumption shows the double-buffered DMA transfers perfectly overlapped with computation to hide their latency.
GAP8 and GAP9 exhibit similar behavior, albeit at significantly different time scales: soft-float emulation on GAP8, due to the lack of FPUs, introduces more than $10\times$ overhead.
As a result, in the \textit{peak throughput} configuration of GAP8, one epoch takes \SI{1311}{\milli\second} vs \SI{102}{\milli\second} on GAP9.
GAP9 also more than halves the average power consumption from \SI{102}{\milli\watt} to \SI{38}{\milli\watt}, which compounds to reduce the energy required for fine-tuning by $34\times$.
The \textit{most energy efficient} configuration captures the case in which conserving energy is more important than the fine-tuning latency (\eg when the drone lands to perform the fine-tuning process).
In this operating point, GAP9 reduces energy consumption by $-25\%$ compared to peak throughput while increasing fine-tuning latency by $53\%$.
These results confirm the feasibility of on-device learning aboard ultra-low power embedded devices in a real-world robotic application.

\rebuttal{
Figure~\ref{fig:impl-optimization} measures the impact of the software optimizations described in Section~\ref{sec:method_impl} on our kernels' performance for the \trainfc{} fine-tuning strategy.
Combined, the four optimizations achieve an $80\times$ speed-up compared to a baseline platform-agnostic C implementation.
Specifically, double buffering hides the latency of L3 DMA transfers by overlapping them with computation.
Albeit this has limited impact on the baseline implementation, DMA transfers impose a fixed \SI{20}{\milli\second} cost that is not impacted by other optimizations, \ie they would account for more than 50\% of total latency of a fully optimized implementation.
Loop unrolling, on the other hand, hides load stalls due to L2 accesses. Four-iteration unrolling on the output tensor dimension (\ie a fully-unrolled kernel, as our model has four-element outputs for the \trainfc{} strategy), yields a $-30\%$ improvement.
The most impactful improvement, more than $9\times$, is achieved by exploiting L1 memory for all buffers in the kernels' inner loops, with L2-L1 transfers handled by DMA.
Finally, parallelization yields a $5\times$ improvement by distributing work among our chip's 8 cores (\ie a 60\% scaling efficiency).}
\section{Conclusion} \label{sec:conclusion}

We present on-device learning aboard nano-drones to mitigate the general TinyML problem of domain shift.
Our fine-tuning approach requires only \SI{19}{\milli\watt}, \SI{1}{\mega\byte} of memory, and runs in just \SI{510}{\milli\second} (5 epochs) on the best-in-class GWT GAP9 SoC.
We employ self-supervised learning to cope with the lack of ground-truth labels aboard our UAV.
In-field results show an improvement in control performance up to 26\% vs. a non-fine-tuned SotA baseline, making the difference between mission failure and success in a never-seen-before challenging environment.

\ifCLASSOPTIONcaptionsoff
  \newpage
\fi


\bibliographystyle{IEEEtran}
\bibliography{bibliography}


\newpage

\begin{IEEEbiography}[{\includegraphics[width=1in,height=1.25in,clip,keepaspectratio]{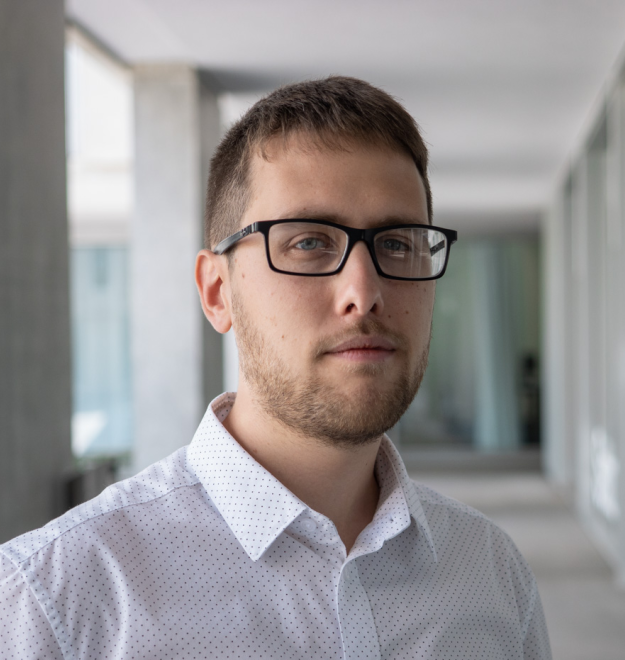}}]{Elia Cereda}
is a third-year Ph.D. student at the Dalle Molle Institute for Artificial Intelligence (IDSIA, USI-SUPSI) in Lugano, Switzerland. He received a double MSc degree from the University of Milano-Bicocca and USI, Lugano in 2021. His research focuses on developing AI-based autonomous algorithms for pocket-sized robotic platforms. He received Best Paper Awards at the IEEE ICCE'18 conference and ACM EWSN'23 SPICES workshop and was a member of the winning team of the first ``Nanocopter AI Challenge'' hosted at the IMAV'22 International Conference.
\end{IEEEbiography}

\begin{IEEEbiography}[{\includegraphics[width=1in,height=1.25in,clip,keepaspectratio]{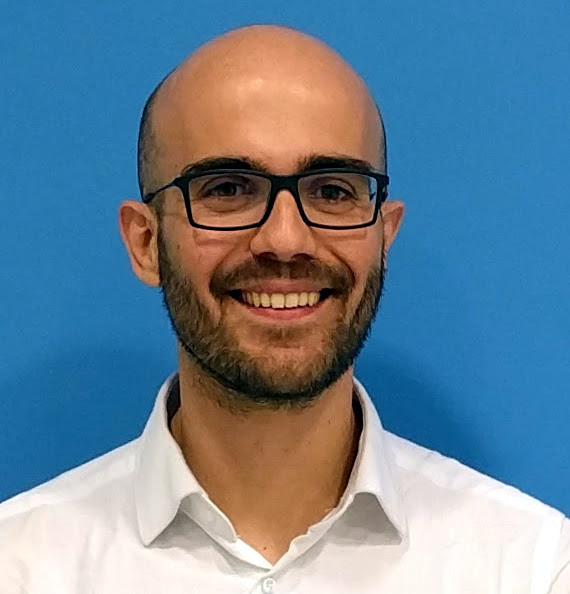}}]{Alessandro Giusti} is Professor of AI for Autonomous Robotics at the Dalle Molle Institute for Artificial Intelligence (IDSIA, USI-SUPSI) in Lugano, Switzerland. He leads the institute's research area on Autonomous Robotics. His work focuses on Self-Supervised Deep Learning applied to mobile and industrial robotics. He is the author of more than 80 peer-reviewed publications in top conferences and journals and the recipient of several awards, most of which for innovative applications of deep learning to various fields.
\end{IEEEbiography}

\begin{IEEEbiography}[{\includegraphics[width=1in,height=1.25in,clip,keepaspectratio]{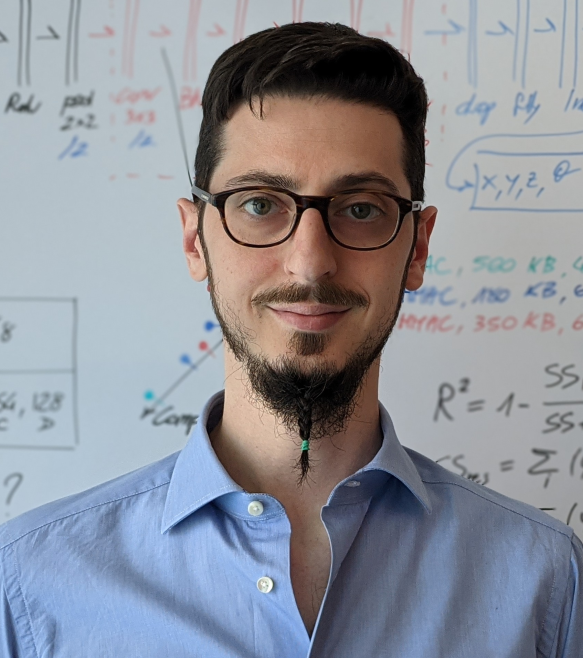}}]{Daniele Palossi}
(he/his) received his Ph.D. in Information Technology and Electrical Engineering from ETH Z\"urich. He is currently a Senior Researcher at the Dalle Molle Institute for Artificial Intelligence (IDSIA), USI-SUPSI, Lugano, Switzerland, where he leads the nano-robotics research group, and at the Integrated Systems Laboratory (IIS), ETH Z\"urich, Z\"urich, Switzerland. His research stands at the intersection of artificial intelligence, ultra-low-power embedded systems, and miniaturized robotics. His work has resulted in 45+ peer-reviewed publications in international conferences and journals. Dr. Palossi was a recipient of the Swiss National Science Foundation (SNSF) Spark Grant, the 2nd prize at the Design Contest held at the ACM/IEEE ISLPED'19, several Best Paper Awards, and team leader of the winning team of the first ``Nanocopter AI Challenge'' hosted at the IMAV'22 International Conference.
\end{IEEEbiography}

\vfill


% \cleardoublepage
% \input{07-rebuttal}

\end{document}